%% file: main.tex
\documentclass[acmtog,nonacm]{acmart} %

\usepackage{booktabs} %

\usepackage{tabularx} %
\usepackage{xspace} %
\usepackage{xcolor} %
\usepackage{makecell} %
\usepackage{multirow} %
\usepackage{bm} %
\usepackage{mathtools}  %
\usepackage{colortbl} %
\usepackage{subcaption}  %

\newcommand{\OURS}{NeRSemble}
\newcommand{\HG}{\mathcal{H}} %
\newcommand{\D}{\mathcal{D}} %
\newcommand{\z}{\bm{\omega}} %

\newcommand{\x}{\mathbf{x}} %

\makeatletter
\DeclareRobustCommand\onedot{\futurelet\@let@token\@onedot}
\def\@onedot{\ifx\@let@token.\else.\null\fi\xspace}

\def\wrt{w.r.t\onedot} 

\makeatother

\newcommand*\diff{\mathop{}\!\mathrm{d}}  %

\newcolumntype{Y}{>{\centering\arraybackslash}X}
\newcolumntype{P}[1]{>{\centering\arraybackslash}p{#1}}
\newcolumntype{C}{>{\centering}X}
\newcommand{\nparticipants}{222}
\newcommand{\nroughparticipants}{220}
\newcommand{\nmaleparticipants}{157}
\newcommand{\nfemaleparticipants}{65}
\newcommand{\nsequences}{4734}
\newcommand{\nroughsequences}{4700}
\newcommand{\nframes}{31.7 million}
\newcommand{\totalrecordingtime}{7h 30m}
\newcommand{\totaldiskspace}{203 TB}

\acmJournal{TOG}
\acmYear{2023} 
\acmVolume{42} 
\acmNumber{4} 
\acmArticle{} 
\acmMonth{8} 
\acmPrice{15.00}

\setcopyright{acmlicensed}

\acmDOI{10.1145/3592455}

\begin{document}
\title{\OURS: Multi-view Radiance Field Reconstruction of Human Heads}

\author{Tobias Kirschstein}
\orcid{0009-0002-5308-591X}
\affiliation{%
  \institution{Technical University of Munich}
  \country{Germany}
}
\email{tobias.kirschstein@tum.de}

\author{Shenhan Qian}
\orcid{0000-0003-0416-7548}
\affiliation{%
  \institution{Technical University of Munich}
  \country{Germany}
}
\email{shenhan.qian@tum.de}

\author{Simon Giebenhain}
\orcid{0000-0002-7588-8767}
\affiliation{%
  \institution{Technical University of Munich}
  \country{Germany}
}
\email{simon.giebenhain@tum.de}

\author{Tim Walter}
\orcid{0009-0008-6618-1824}
\affiliation{%
  \institution{Technical University of Munich}
  \country{Germany}
}
\email{tim.michelbach@hotmail.com}

\author{Matthias Nießner}
\orcid{0000-0001-6093-5199}
\affiliation{%
  \institution{Technical University of Munich}
  \country{Germany}
}
\email{niessner@tum.de}

\begin{teaserfigure}
  \includegraphics[width=\textwidth]{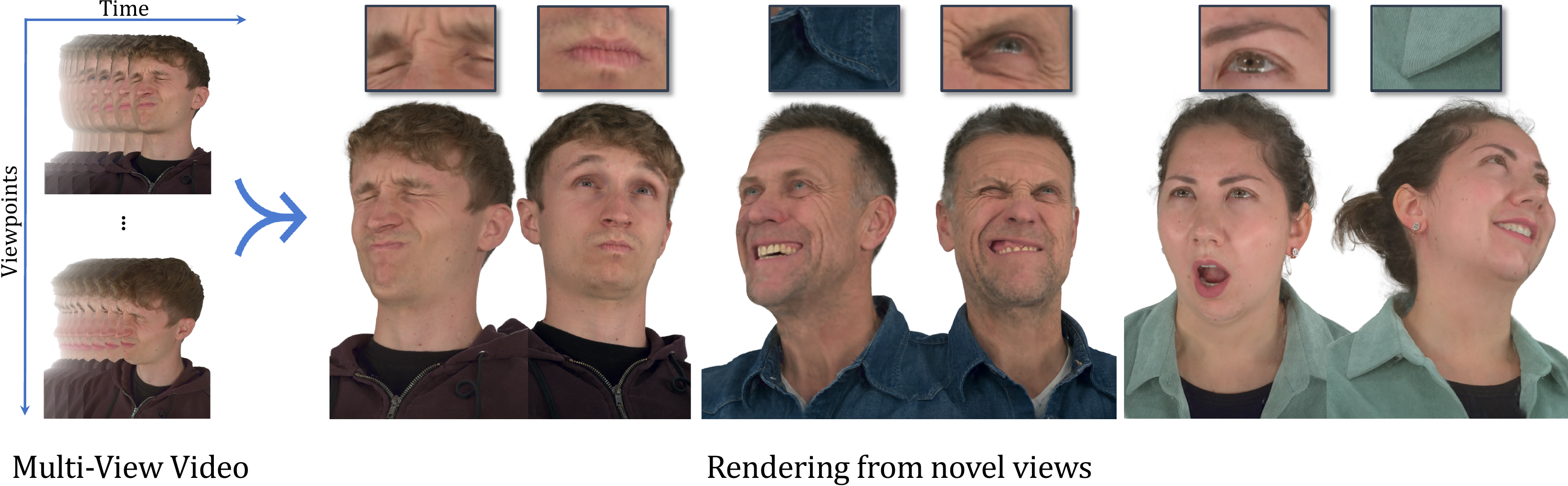}
  \caption{\textbf{\OURS:} Given multi-view video recordings from twelve cameras (left), our method is capable of synthesizing highly realistic novel views of human heads in complex motion.
  Our renderings from unseen views (right) faithfully represent static scene parts and regions undergoing highly non-rigid deformations.
  Along with our method, we publish our high-quality multi-view video capture data of {\nframes} frames from a total of {\nparticipants} subjects.
  }
  \label{fig:teaser}
\end{teaserfigure}

\input{include/00_abstract}

\ccsdesc[300]{Computing methodologies~Rendering}
\ccsdesc[300]{Computing methodologies~3D imaging}
\ccsdesc[300]{Computing methodologies~Volumetric models}
\ccsdesc[300]{Computing methodologies~Reconstruction}

\keywords{
Neural Radiance Fields, Dynamic Scene Representations, Novel View Synthesis, Multi-View Video Dataset, Human Heads
}

\maketitle

\input{include/01_introduction}
\input{include/02_related_work}

\input{include/03_dataset}

\input{include/04_method}

\input{include/05_experimental_results}
\input{include/06_conclusion.tex}

\begin{acks}
This work was supported by the ERC Starting Grant Scan2CAD (804724), the German Research Foundation
(DFG) Grant “Making Machine Learning on Static and Dynamic 3D Data Practical”, and the German Research Foundation (DFG) Research Unit “Learning and Simulation in Visual Computing”. We would also like to thank Maximilian Knörl for the help with data acquisition, and Angela Dai for the video voice-over.
\end{acks}

\bibliographystyle{ACM-Reference-Format}
\bibliography{bibliography}

\end{document}

%% file: include/00_abstract.tex
\begin{abstract}

We focus on reconstructing high-fidelity radiance fields of human heads, capturing their animations over time, and synthesizing re-renderings from novel viewpoints at arbitrary time steps.
To this end, we propose a new multi-view capture setup composed of 16 calibrated machine vision cameras that record time-synchronized images at 7.1~MP resolution and 73 frames per second.
With our setup, we collect a new dataset of over {\nroughsequences} high-resolution, high-framerate sequences of more than {\nroughparticipants} human heads, from which we introduce a new human head reconstruction benchmark\footnote{We will release all of our captured data, including all {\nsequences} recordings and baseline codes, along with a new public benchmark to support further research in the area.}.
The recorded sequences cover a wide range of facial dynamics, including head motions, natural expressions, emotions, and spoken language.
In order to reconstruct high-fidelity human heads, we propose Dynamic Neural Radiance Fields using Hash Ensembles ({\OURS}). We represent scene dynamics by combining a deformation field and an ensemble of 3D multi-resolution hash encodings. The deformation field allows for precise modeling of simple scene movements, while the ensemble of hash encodings helps to represent complex dynamics. 
As a result, we obtain radiance field representations of human heads that capture motion over time and facilitate re-rendering of arbitrary novel viewpoints.
In a series of experiments, we explore the design choices of our method and demonstrate that our approach outperforms state-of-the-art dynamic radiance field approaches by a significant margin.

{\let\thefootnote\relax\footnotetext{\footnotesize{Website: {\color[HTML]{0065bd}\href{https://tobias-kirschstein.github.io/nersemble}{\texttt{https://tobias-kirschstein.github.io/nersemble}}}}}}
\end{abstract}

%% file: include/01_introduction.tex
\section{Introduction}
\label{sec:intro}

In recent years, we have seen tremendous growth in the importance of digital applications that rely on photo-realistic rendering of images from captured scene representations, both in society and industry.
In particular, the synthesis of novel views of dynamic human faces and heads has become the center of attention in many graphics applications ranging from computer games and movie productions to settings in virtual or augmented reality. 
Here, the key task is the following: given a recording of a human actor who is displaying facial expressions or talking, reconstruct a temporally-consistent 3D representation.
This representation should enable the synthesis of photo-realistic re-renderings of the human face from arbitrary viewpoints and time steps.

However, reconstructing a 3D representation capable of photo-realistic novel viewpoint rendering is particularly challenging for dynamic objects.
Here, we not only have to reconstruct the static appearance of a person, but we also have to simultaneously capture the motion over time and encode it in a compact scene representation.
The task becomes even more challenging in the context of human faces, as fine-scale and high-fidelity detail are required for downstream applications, where the tolerance for visual artifacts is typically very low. 
In particular, human heads exhibit several properties that make novel view synthesis (NVS) extremely challenging, such as the complexity of hair, differences in reflectance properties, and the elasticity of human skin that creates heavily non-rigid deformations and fine-scale wrinkles.

In the context of static scenes, we have seen neural radiance field representations (NeRFs)~\cite{mildenhall2020nerf} obtain compelling NVS results.
The core idea of this seminal work is to leverage a volumetric rendering formulation as a reconstruction loss and encode the resulting radiance field in a neural field-based representation. 
Recently, there has been significant research interest in extending NeRFs to represent dynamic scenes.
While some approaches rely on deformation fields to model dynamically changing scene content \cite{park2021nerfies,park2021hypernerf}, others propose to replace the deformation field in favor of a time-conditioned latent code \cite{li2022dynerf}. 
These methods have shown convincing results on short sequences with limited motion; however, faithful reconstructions of human heads with complex motion remain challenging.

In this work, we focus on addressing these challenges in the context of a newly-designed multi-view capture setup and propose \OURS, a novel method that combines the strengths of deformation fields and flexible latent conditioning to represent the appearance of dynamic human heads.
The core idea of our approach is to store latent features in an ensemble of multi-resolution hash grids, similar to Instant~NGP~\cite{mueller2022instant}, which are blended to describe a given time step. 
Importantly, we utilize a deformation field before querying features from the hash grids.
As a result, the deformation field represents all coarse dynamics of the scene and aligns the coordinate systems of the hash grids, which are then responsible for modeling fine details and complex movements. 
In order to train and evaluate our method, we design a new multi-view capture setup to record 7.1~MP videos at 73~fps with 16 machine vision cameras. 
With this setup, we capture a new dataset of {\nsequences} sequences of {\nparticipants} human heads with a total of {\nframes} individual frames.
We evaluate our method on this newly-introduced dataset and demonstrate that we significantly outperform existing dynamic NeRF reconstruction approaches.
Our dataset exceeds all comparable datasets \wrt resolution and number of frames per second by a large margin, and will be made publicly available.
Furthermore, we will host a public benchmark on dynamic NVS of human heads, which will help to advance the field and increase comparability across methods.

\smallskip
To summarize, our contributions are as follows:
\begin{itemize}
    \item A dynamic head reconstruction method based on a NeRF representation that combines a deformation field and an ensemble of multi-resolution hash encodings. This facilitates high-fidelity NVS from a sparse camera array and enables detailed representation of scenes with complex motion.
    \item A high-framerate and high-resolution multi-view video \\dataset of diverse human heads with over {\nroughsequences} sequences of more than {\nroughparticipants} subjects. The dataset will be publicly released and include a new benchmark for dynamic NVS of human heads.
\end{itemize}

\input{tables/existing_datasets}

%% file: tables/existing_datasets.tex
\begin{table}
    \small
    \renewcommand{\arraystretch}{1.2}
    \centering

    \caption{Existing multi-view video datasets of human faces. Note that for each dataset, we only count the publicly accessible recordings.}
    
    \begin{tabularx}{\linewidth}{lrrrr}
        \toprule
        \addlinespace[-2pt]
        Dataset & \#Subj. & \#Cam.  & Resolution & Fps \\
        \midrule

        D3DFACS~{[\citeyear{cosker2011facs}]} & 10 & 6 & 1280 x 1024 & 60 \\
        BP4D-Spontaneous~{[\citeyear{zhang2014bp4d}]} & 41 & 3 & 1392 x 1040 & 25 \\
        Interdigital Light-Field~{[\citeyear{Sabater2017}]} & 5 & 16 & 2048 x 1088 &  30 \\
        4DFAB~{[\citeyear{cheng20184dfab}]} & 180 & 7 & 1600 x 1200 & 60 \\
        VOCASET~{[\citeyear{VOCA2019}]} & 12 & 12 & 1600 x 1200 & 60 \\
        MEAD~{[\citeyear{kaisiyuan2020mead}]} & 48 & 7 & 1920 x 1080 &  30\\
        MultiFace~{[\citeyear{wuu2022multiface}]} & 13 & 150 & 2048 x 1334 & 30\\
        \midrule 
        \addlinespace[-1pt]
        Ours & \nparticipants & 16 & 3208 x 2200 & 73 \\
        \bottomrule
    \end{tabularx}
    
    \label{tab:03_existing_datasets}
\end{table}

%% file: include/02_related_work.tex
\begin{figure*}[tb]
    \centering
    \includegraphics[width=\linewidth]{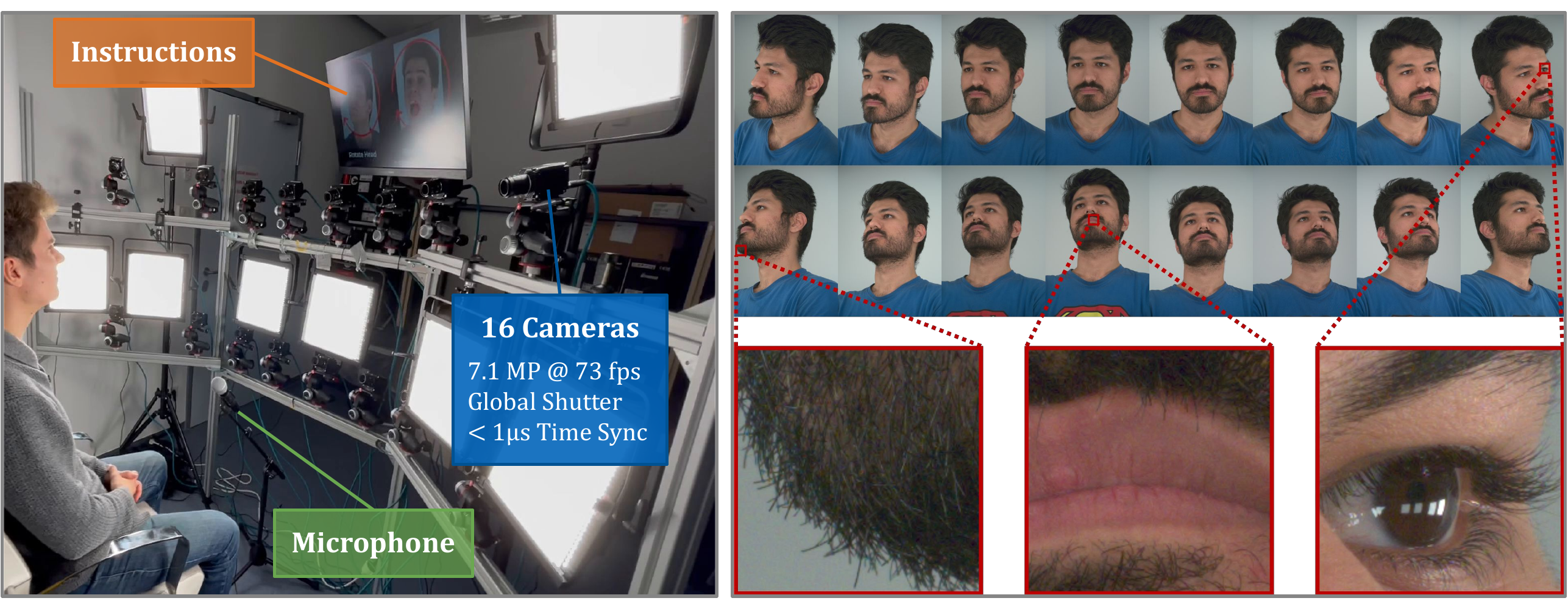}
    \caption{Left: Our custom-built multi-view video capture setup. Right: The 16 viewpoints and the facial detail obtained from the recordings.}
    \label{fig:03_capture_setup}
\end{figure*}

\section{Related Work}
\label{sec:related_work}

Modeling and rendering human faces is a central topic in graphics and plays a crucial role in many applications, such as computer games, social media, telecommunication, and virtual reality.

\subsection{3D Morphable Models}
3D morphable models (3DMMs) have been a staple approach over the last two decades. The use of a unified mesh topology enables representing identity and expression using simple statistical tools~\cite{blanz_vetter, FLAME:SiggraphAsia2017}. With the additional use of texture, one can already produce compelling renderings~\cite{blanz_vetter, BFM}, but mesh-based 3DMMs are inherently limited w.r.t. modeling hair or fine identity-specific details.
More recently, the use of neural fields~\cite{neural_fields_survey} has alleviated the constraint of working on topologically uniform meshes. These models are capable of modeling complete human heads, including hair \cite{yenamandra2021i3dmm} and fine details \cite{giebenhain2022nphm}.
In another line of work, \citet{zheng2022imavatar} combine ideas from neural fields and classical 3DMMs to fit monocular videos.

 \subsection{Neural Radiance Fields}
Our work strives to achieve highly-realistic renderings of videos, including detailed hairstyles and complex deformations. Therefore, we deviate from common assumptions made in 3DMMs and focus on fitting a single multi-view video sequence to the highest degree of detail possible.
Neural Radiance Fields (NeRFs)~\cite{mildenhall2020nerf} have recently become state-of-the-art in NVS.
While the first NeRFs were usually trained for hours or days on a single scene, recent research advances have reduced the training time to several minutes.
For example, this can be achieved by grid-based optimization~\cite{yu_and_fridovichkeil2021plenoxels, SunSC22dvgo, karnewar2022relu}, tensor decomposition~\cite{tensoRF}, or Instant~NGP's~\cite{mueller2022instant} multi-resolution voxel hashing.

\subsection{Dynamic NeRF}
Extending NeRFs to time-varying, non-rigid content is another central research topic that has seen fast progress.
\citet{D-NeRF} and Park et al. \shortcite{park2021nerfies, park2021hypernerf} model a single NeRF in canonical space and explicitly model backward deformations from observed frames to explain the non-rigid content of the scene. 
OLD: On the other hand, \citet{li2022dynerf} refrain from using explicit deformations and instead encode the state of the scene in a latent vector, which is directly conditioning a NeRF.
\citet{FourierPlenOctress} utilize Fourier-based compression of grid features to represent a 4D radiance field. \citet{NeuralVolumes} use an image-to-volume generator in conjunction with deformation fields.

Concurrent to our work, \citet{song2022nerfplayer} combine a fast NeRF backbone, i.e. TensoRF or Instant~NGP, with a sliding window approach to account for temporal changes. 
\citet{attal2023hyperreel} combine a 4D tensor decomposition with a learned sampling method for fast dynamic NVS. 
In contrast to these works, we propose a hash-based decomposition in conjunction with a deformation field. 

\subsection{Video View Synthesis}
Besides NeRF, there also exist other methods for video view synthesis that do not rely on a radiance field backbone.
In an early work, \citet{zitnick2004high} use geometry-assisted image-based rendering to render novel views of dynamic scenes. %
More recently, \citet{broxton2020immersive} obtain free viewpoint videos by constructing multi-sphere images that are then transformed into a layered mesh representation for fast rendering and streaming.
A different approach is pursued by \citet{collet2015high}, who obtain tracked meshes of dynamic performances with a multi-view stereo system. 
While these mesh-based methods produce compelling video view synthesis for larger scenes, the strength of NeRFs lies in photo-realistic reconstruction of fine and complex details such as hair.

\subsection{NeRF for Faces}
Several works propose methods specialized to the domain of human heads. 
Notably, \citet{Gafni_2021_CVPR_nerface} use fitted 3DMM parameters to condition a NeRF, and \citet{RigNeRF} extend this approach to model explicit deformations derived from the 3DMM's geometry. 
More recently, \citet{zielonka2022insta} propose a similar approach focused on reconstruction speed and real-time rendering by utilizing a tracked 3DMM in conjunction with Instant~NGP.
\citet{MoRF} propose a generative NeRF with control over identity parameters. \citet{hong2021headnerf} pursue a similar approach with additional expression parameters.
\citet{lombardi2021mixture} propose a highly-optimized approach to neural rendering by explicitly storing color emission values in voxel grids that are loosely rigged to a 3DMM's surface. 
In this work, we propose a template-free approach as we argue that it is difficult to achieve pixel-accurate novel view synthesis with coarse geometry proxies such as FLAME \cite{FLAME:SiggraphAsia2017}.

Similar to our method, \citet{Gao2022nerfblendshape} recently proposed to blend features from multiple hash grids. While their approach uses parameters from a tracked 3DMM, \OURS~jointly optimizes for blend weights and the remaining model parameters. Additionally, we show that including a deformation field before blending the hash grids brings significant improvements.

%% file: include/03_dataset.tex
\section{Multi-view video dataset of human faces}

We introduce a novel dataset consisting of {\nsequences} multi-view video recordings of {\nparticipants} subjects that were captured with 16 machine vision cameras.
Our forward-facing capture rig covers a field of view of 93° left-to-right and 32° up-to-down.
As human face motion is complex and the perceived emotion can be heavily influenced by subtle differences, we use a high resolution of 7.1 megapixels, encompassing the whole face up to the level of individual hair strands and wrinkles, as shown in Figure~\ref{fig:03_capture_setup}.
We also ensure that no subtle movements are missed by recording at 73 frames per second.
Taken together, our dataset is a unique combination of high-resolution, high frame-rate recordings of many subjects, which is currently unmatched by any other dataset (see Table~\ref{tab:03_existing_datasets}).

\input{tables/dataset_statistics}%
\begin{figure}[tb]
    \vspace{-0.05cm}  %
    \centering
    \includegraphics[width=\linewidth]{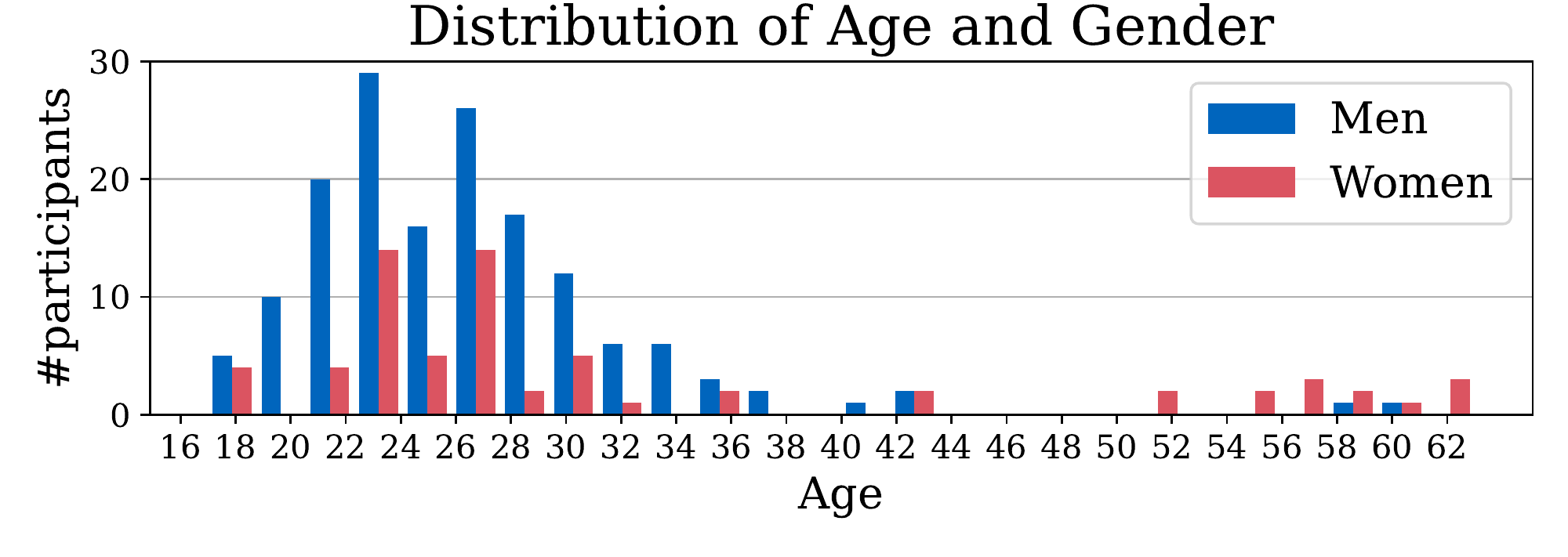}
    \raggedleft 
    \includegraphics[width=0.95\linewidth]
    {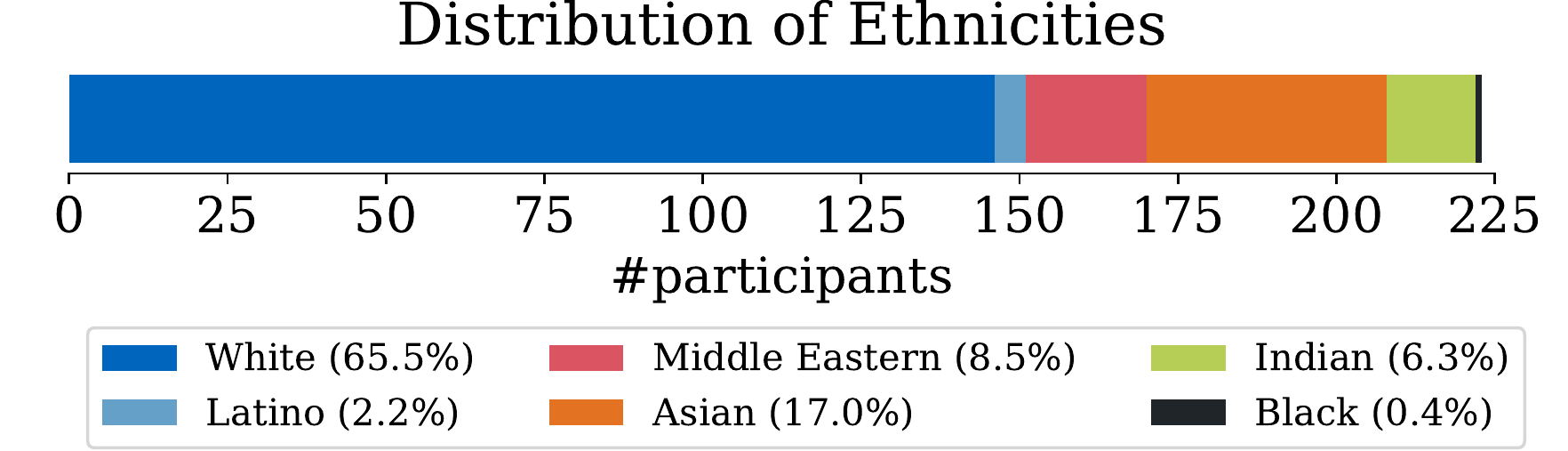}
    \caption{\textbf{Statistics of the participants in our dataset.} Our recorded sequences feature a wide range of ages and ethnicities from both genders.}
    \label{fig:dataset_statistics}
\end{figure}

\subsection{Acquisition}
Each recording session consists of 25 short sequences, resulting in around 3 minutes of multi-view video footage per person. 
We ask the participants to perform a diverse set of facial expressions in order to maximize the variety of motion. 
Specifically, our capture script consists of 9 expression sequences covering different facial muscle groups, 1 hair sequence with fast movements, 4 emotion sequences, 10 sentences with audio, and 1 longer sequence where subjects are free to perform arbitrary facial deformations and head motions.

To obtain high-quality video recordings, we employ a shutter speed of 3ms, which allows us to capture fast movements while avoiding motion blur. 
Furthermore, we use a small lens aperture to obtain sharp images everywhere in the face region. 
This combination yields high-quality captures but reduces the amount of incident light at the camera sensors, which requires us to illuminate our scene with 8 strong LED light panels.
We further use diffusor plates on the lights to reduce specularities on the skin.
Additionally, our cameras employ the precision time protocol (PTP) for accurate time synchronization. The synchronized clocks have sub-microsecond accuracy, resulting in video frames that are effectively captured simultaneously.
Finally, we make use of a color checker to calibrate the white balancing factors as well as the gamma parameters of each camera. The resulting video recordings have consistent colors across viewpoints and capture fine details as shown in Figure~\ref{fig:03_capture_setup}.

\begin{figure}
    \centering
    \includegraphics[width=\linewidth]{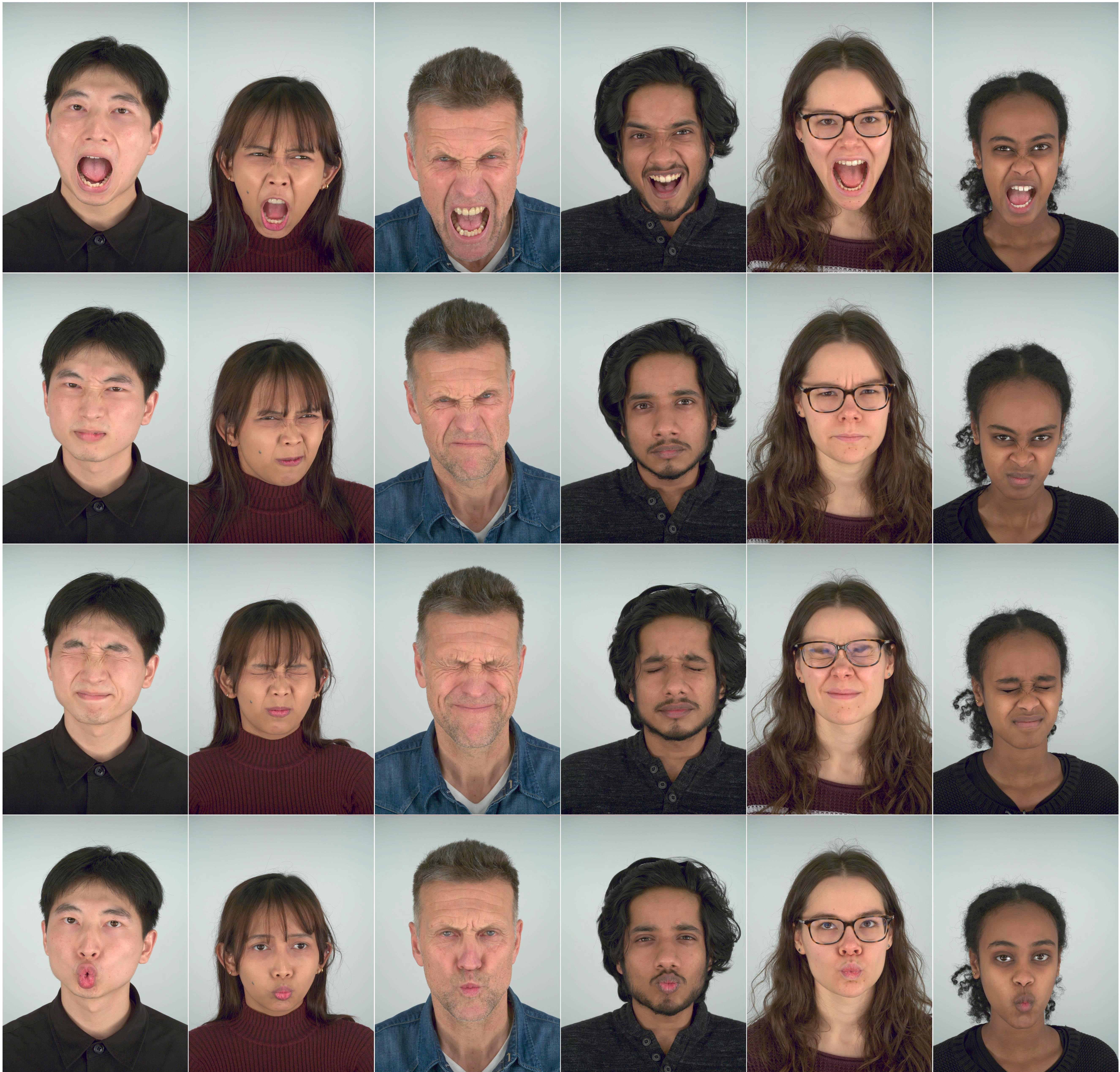}
    \caption{\textbf{Structure of our dataset.} We ask every participant to perform the same sequence of expressions.}
    \label{fig:my_label}
\end{figure}

\subsection{Processing}

\begin{figure*}[t]
    \centering
    \includegraphics[width=\linewidth]{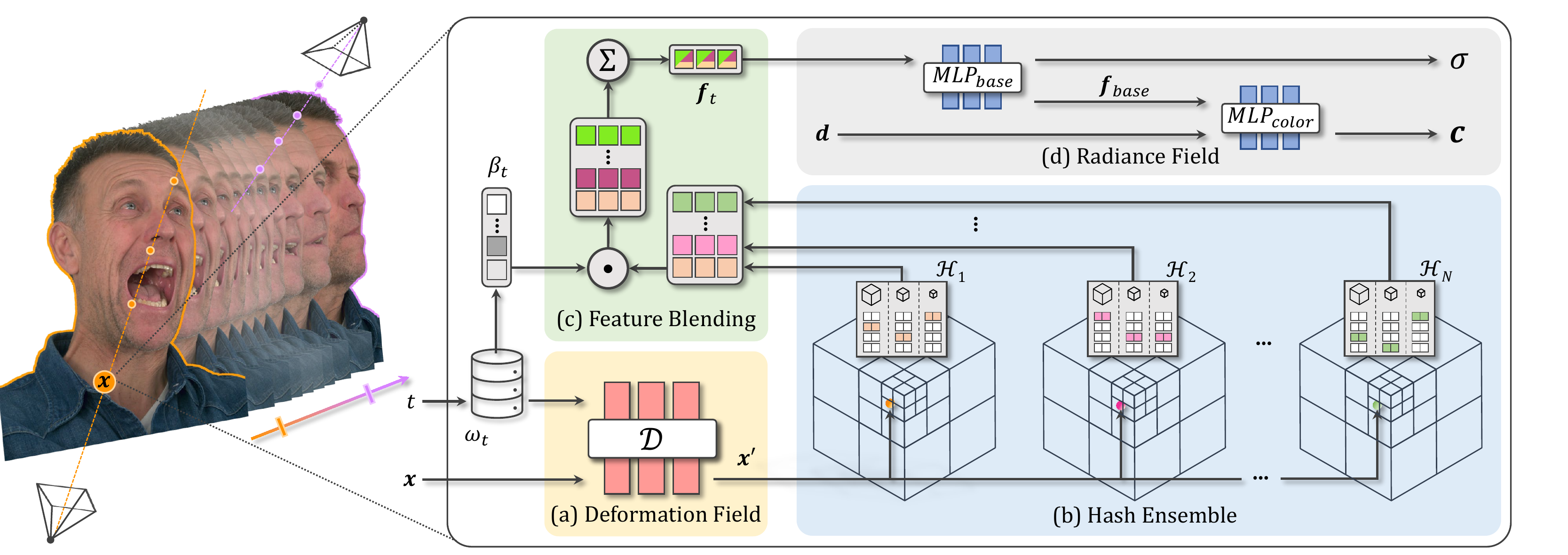}
    \caption{\textbf{Method Overview.} 
    \OURS~represents a spatio-temporal radiance field for dynamic NVS using volume rendering (left).
    On the right side, we show how \OURS~obtains a density $\sigma(\x)$ and color value $\mathbf{c}(\x, \mathbf{d})$ for a point $\x$ on a ray at time $t$.
    (a) Given the deformation code $\z_t$ the point $\x$ is warped to $\x' = \D(\x, \z_t)$ in the canonical space. 
    (b) The resulting point is used to query features $\HG_i(\x')$ from the $i$-th hash grid in our ensemble. 
    (c) The resulting features are blended using weights $\beta_t$. Note that both $\z_t$ and $\beta_t$ contribute to explaining temporal changes.
    (d) We predict density $\sigma(\x)$ and view-dependent color $\mathbf{c}(\x, \mathbf{d})$ from the blended features using an efficient rendering head consisting of two small MLPs.
    }
    \label{fig:method_overview}
\end{figure*}

We estimate an individual extrinsic and a shared intrinsic camera matrix by employing a fine checkerboard in combination with a bundle adjustment optimization procedure. This leads to accurate estimated camera poses, which we verified to be in the sub-millimeter regime in a synthetic setting.
Furthermore, the background of our recordings is a white wall, which is captured prior to recording. From these empty backgrounds, it is later feasible to obtain high-quality foreground segmentation maps for each frame via image matting methods, e.g., using BackgroundMatting v2 \cite{BGMv2}. 

\subsection{Benchmark}

Our dataset enables us to study photo-realistic human head reconstruction from multi-view videos, which is the goal of this work. 
Moreover, the captured data allows for use cases far beyond NVS such as generalization over human heads, immersive video conferencing, VR-ready avatar rendering, studying microexpressions, re-enacting, animating, and many more. 
As such, we plan to release the full dataset to the academic community.
Furthermore, we will use a representative selection of recordings from our dataset to compile a benchmark for NVS on human faces. 
We hope that this endeavor promotes comparability across methods and ultimately advances research on high-fidelity human head reconstruction.

\subsection{Data Privacy}
Due to the sensitivity of the captured data, all participants in our dataset signed an agreement form compliant with GDPR requirements. Please note that GDPR compliance includes the right for every participant to request the timely deletion of their data. We will enforce these rights in the distribution of our dataset.

%% file: tables/dataset_statistics.tex
\begin{table}[tb]
    \centering
    \small
    \caption{\textbf{Statistics of our multi-view video face dataset.}}
    
    \begin{tabularx}{1\linewidth}{ccccc}
        \toprule
        \#Participants &
        \#Sequences &
        \#Frames &
        Total Time &
        Disk Space \\
        \midrule
        {\nparticipants} (\nmaleparticipants m / \nfemaleparticipants f) &
        \nsequences &
        \nframes &
        \totalrecordingtime &
        \totaldiskspace\\
        \bottomrule
    \end{tabularx}
\end{table}

%% file: include/04_method.tex
\section{Dynamic NeRF using Hash Ensembles}
\label{sec:method}

Our goal is to find a spatio-temporal representation that allows for highly realistic NVS of human heads undergoing complex non-rigid deformations. 
To this end, we propose a combination of a deformation field and a decomposition of the 4D scene volume into an ensemble of 3D feature volumes along the temporal dimensions in order to reconstruct the dynamics of a scene (see Figure \ref{fig:method_overview}).

\subsection{Preliminaries: Neural Radiance Fields}
\label{sec:preliminaries}

Our work builds on top of the recent success of Neural Radiance Fields (NeRFs)~\cite{mildenhall2020nerf}, which utilize volume rendering through a density field $\sigma(\mathbf{x})$ and view-dependent color field $\mathbf{c}(\mathbf{x}, \mathbf{d})$. 
Given a ray $\mathbf{r}(\tau) = \mathbf{o} + \tau\mathbf{d}$, a color value 
\begin{equation}
    C(\mathbf{r}) = \int_{\tau_n}^{\tau_f} \underbrace{T(\tau)\sigma(\mathbf{r}(\tau))}_{w(\tau)}\mathbf{c}(\mathbf{r}(\tau), \mathbf{d}) \diff \tau,
    \label{eq:volume_rendering}
\end{equation}
is obtained by integrating from near plane $\tau_n$ to far plane $\tau_f$ along the ray, where $T(\tau) = \text{exp}\left(-\int_{\tau_n}^{\tau}\sigma(\mathbf{r}(s))\diff s\right)$ denotes the accumulated transmittance up to $\tau$.

The goal of the optimization is to solve for the optimal parameters of a multilayer perceptron (MLP) that encode the resulting radiance field.
Recently, pure voxel grids~\cite{yu_and_fridovichkeil2021plenoxels} and combinations of explicit grids with MLPs~\cite{mueller2022instant} have been demonstrated to be effective alternatives for the radiance field representation.

\paragraph{Instant~NGP}
Our method relies on the voxel hashing scheme of Instant~NGP~\cite{mueller2022instant}, which uses multi-resolution features $\mathbf{f}(\x)$ in combination with two small MLPs to represent the 3D fields of a NeRF:
\begin{align}
    [\sigma(\x), \mathbf{f}_{\text{base}}(\x)] &= \text{MLP}_{\text{base}}(\mathbf{f}(\x)) \\
    \mathbf{c}(\x, \mathbf{d}) &= \text{MLP}_{\text{color}}(\mathbf{f}_{\text{base}}(\x), \mathbf{d}).
    \label{eq:base_MLP}
\end{align}
Importantly, the features are stored in a multi-resolution hash grid $\HG$, s.t. $f(\x)=\HG(\x)$. The hash grid $\HG$ provides a memory-efficient way to encode the 3D scene volume to a stage where a tiny MLP is powerful enough to represent even the most complex of scenes.

\subsection{Multi-Resolution Hash Ensemble}
\label{sec:hash_ensemble}

Our representation of a dynamic scene is inspired by classical blend shapes \cite{blendshapes}. 
We assume that any state of the scene at time $t$ can be expressed as a combination of feature vectors drawn from a set of multi-resolution hash grids $\{\HG_i\}_{i=1}^N$, which we refer to as an ensemble of hash grids.
To obtain a blended radiance field at time $t$, we formulate it as a linear combination of its features
\begin{equation}
    \mathbf{f}_{t}(\x) = \sum_{i=1}^N \beta_{t, i}\HG_i(\x),
    \label{eq:feature_blending}
\end{equation}
using blend weights $\beta$, which are optimized alongside the hash ensemble, the shared $\text{MLP}_{\text{base}}$ and $\text{MLP}_{\text{color}}$.

This blending operation allows the model to represent complex movements since the blending takes place in feature space. Subsequently, the blended features are decoded by $\text{MLP}_{\text{base}}$ and $\text{MLP}_{\text{color}}$.

\subsection{Spatial Alignment of Features}
\label{sec:alignment}
The blending of hash grid features is most effective if all individual elements of the ensemble are operating in a shared coordinate system. 
For instance, traditional blend shapes operate under perfect correspondences given by the vertex ordering of the mesh topology.  
Since we blend features without such a structure, we explicitly model the deformation using an $SE(3)$ field, represented by a coordinate-based MLP, following \citet{park2021nerfies}. 
More specifically, our deformation field
\begin{equation}
    \D: \mathbb{R}^3 \times \mathbb{R}^{d_{\text{def}}} \rightarrow \mathbb{R}^3, \left(\x, \z_{t}\right) \mapsto \x'
\end{equation}
maps a point $\x$ from observed space to its corresponding point $\x'$ in the canonical space, given the conditioning code $\z_{t}$ which describes the current expression. 
The deformation field then finds corresponding points across time steps and maps them to a shared canonical space. 

Using these learned correspondences, we modify Equation~\ref{eq:feature_blending} to operate in the canonical space:
\begin{equation}
    \mathbf{f}^{(t)} = \sum_{i=1}^N \beta_{t, i} \HG_i\left(\D\left(\x, \z_{t}\right)\right).
    \label{eq:canonical_feature_blending}
\end{equation}
This way it becomes easier to blend features of the same moving point observed at two different timesteps.

\subsection{Warm-Up Phase}
\label{sec:warm_up}
With this combination, the hash ensemble and deformations compete to explain the dynamics of the face. 
Hence, the optimization is likely to result in local minima, in which $\D$ does not provide meaningful deformations. 
Therefore, we propose a warm-up phase in the optimization procedure in order to encourage $\D$ to learn meaningful correspondences between observed and canonical space.

During the first $E_{\text{init}}$ epochs of optimization, we disable all but one hash grid, such that the model essentially mimics a deformable NeRF. 
During this stage, the deformation field $\D$ along with its deformation codes $\z_{t}$ are the only means to explain dynamic behavior. 
Thus, $\D$ is able to learn meaningful deformations undisturbed, which is essential to effective blending of hash table features later on.

After the warm-up, the first hash table along with our deformation field is able to explain low-frequency dynamics of the scene. 
We continue to add all remaining hash tables to the optimization over the course of the next $E_{\text{trans}}$ epochs. 
These successively inserted tables enable us to represent fine-scale motion and detail which otherwise cannot be be represented by $\D$.

In order to ensure a smooth transition, we adapt the blend weights
\begin{equation}
    \beta_{t, i}(s)^{*} = \alpha_i(s)\beta_{t, i} \quad \left( \forall i \in \{1, ..., N\} \right),
\end{equation}
where $i$ indexes the hash ensemble, $\alpha_i(s)$ is the windowing function introduced by \citet{park2021nerfies} and 
$s$ is scheduled to linearly increase from $1$ to $N$ between epochs $E_{\text{init}}$ and $E_{\text{trans}} + E_{\text{init}}$. 
Crucially, $\alpha_1(s)=1$ throughout the complete optimization ensuring that the first hash table is always active.

\subsection{Depth Supervision}
\label{sec:depth_supervision}

Since our multi-view dataset provides the capabilities to compute depth maps via traditional methods, we also study the usefulness of additional depth supervision in this work. 
Given the depth $z^{\mathrm{gt}}(\bm{r})$ of a ray, we compute the depth loss as
\begin{align}
    \mathcal{L}_{\mathrm{depth}} = \mathbb{E}_{\bm{r} \sim \mathcal{R}_{\mathrm{d}}}\left[\left(z(\bm{r}) - z^{\mathrm{gt}}(\bm{r})\right)^2\right],
    \quad  
\end{align}
where the expected depth of ray $\bm{r}$ is  $z(\bm{r}) = \int_{\tau_n}^{\tau_f} w(\tau) \cdot \tau \diff \tau$. \\
Since depth observations are incomplete in practice, the depth loss is only computed on rays $\bm{r} \in \mathcal{R}_{\mathrm{d}}$ for which the depth is known.

Additionally, we adopt the two line-of-sight priors from Urban~Radiance~Fields~(URF)~\cite{rematas2022urf} to further leverage depth constraints. 
First, we utilize 
\begin{equation}
    \mathcal{L}_{\text{empty}} = \mathbb{E}_{\bm{r} \sim \mathcal{R}_{\mathrm{d}}}\left[ \int_{\tau_n}^{z(\bm{r})-\epsilon} w(\tau)^2 d\tau \right]
\end{equation} to carve empty space in front of a surface, where $\epsilon$ is exponentially decayed during training as in URF.
Second, 
\begin{equation}
    \mathcal{L}_{\text{near}} = \mathbb{E}_{\bm{r}\sim \mathcal{R}_{\mathrm{d}}}\left[
    \int_{z(\bm{r})-\epsilon}^{z(\bm{r})+\epsilon} \left(w(\tau) - \mathcal{N}\left(\tau~|~z(\bm{r}), \left(\frac{\epsilon}{3}\right)^2\right)\right)^2d\tau
    \right]
\end{equation}
encourages volumetric density in a neighborhood around the depth observation to follow a narrowing Gaussian distribution.
In conjunction with $\mathcal{L}_{\mathrm{depth}}$, these three priors form the depth supervision that is targeted at improving the geometric fidelity of the reconstruction.

\subsection{Background Removal}

We employ continuous-valued alpha maps $M(\bm{r})$ to discourage the model from reconstructing parts of the background. We use a sparsity enforcing L1 loss that penalizes density on rays that hit background pixels:
\begin{align}
    \mathcal{L}_{\mathrm{mask}} = \mathbb{E}_{\bm{r} \sim \mathcal{R}_{\mathrm{bg}}} \left [\left\| ( 1 - T(\tau_f)) - M(\bm{r}) \right\|_1 \right ],
\end{align}
where $T(\tau_f)$ is the total transmittance of ray $\bm{r}$ and $M(\bm{r})$ is its corresponding alpha value from the precomputed alpha map.

\subsection{Optimization Objective}
\label{sec:optimization}

The final loss is comprised of the following terms:
\begin{align}
    \mathcal{L} &=\mathcal{L}_{\mathrm{rgb}} + \mathcal{L}_{\mathrm{mask}} + \underbrace{\mathcal{L}_{\mathrm{depth}} + \mathcal{L}_{\mathrm{near}} + \mathcal{L}_{\mathrm{empty}}}_{\mathclap{\text{depth supervision}}} + \mathcal{L}_{\mathrm{dist}}
\end{align}
where $\mathcal{L}_{\mathrm{rgb}}$ is the standard MSE color loss, which we only compute on foreground rays.
We also add a distortion loss $\mathcal{L}_{\mathrm{dist}}$, which penalizes isolated islands of low density~\cite{barron2022mipnerf360}.
As our scenes only consist of human heads, which are roughly convex, we further compute $\mathcal{L}_{\mathrm{dist}}$ on random rays pointing towards the center. 
This extends the term's regularization effect to the space behind the head, which is often occluded in our scenario. 

Finally, we equip each loss term with a corresponding weight: %

${\lambda_{\mathrm{depth}}, \lambda_{\mathrm{dist}}, \lambda_{\mathrm{near}}, \lambda_{\mathrm{empty}} = 1\mathrm{e-}4}$ and ${\lambda_{\mathrm{mask}} = 1\mathrm{e-}2}$.

\subsection{Discussion on Dynamic Scene Representations}

\paragraph{Relation to Tensor Decomposition}
Equation~\ref{eq:feature_blending} can be interpreted as a special case of a 4D tensor decomposition, similar to the vector-matrix decomposition introduced in TensoRF \cite{tensoRF}. A spatio-temporal tensor $\mathcal{T} \in \mathbb{R}^{D_T \times D_X \times D_Y \times D_Z}$ representing a dynamic scene can be decomposed into a sum of four vector-tensor outer-products:
\begin{equation}
    \mathcal{T} \approx \sum_{a \in A}\sum_{i=1}^N v^a_i \circ M^{A\setminus \{a\}}_i,
    \label{eq:tensorf}
\end{equation}
where $\circ$ denotes the vector-tensor outer product, $A = \{X, Y, Z, T\}$ is the set of axis indices, $v^a_i \in \mathbb{R}^{D_a}$ is a vector and $M^{A\setminus \{a\}}_i$ is a 3D tensor, for example $M^{A\setminus \{X\}}_i \in \mathbb{R}^{D_T \times D_Y \times D_Z}$.

Equation~\ref{eq:feature_blending} of our method can be seen as a special case of Equation~\ref{eq:tensorf}, where only the term for $a=T$ is used. Instead of storing features in a dense grid $M^{A\setminus \{a\}}_i$, we employ a memory efficient hash table representation $M^{A\setminus \{T\}}_i = \mathcal{H}_i$ and the vector $v_i^a$ corresponds to our blend weights $v_{i,t}^a = \beta_{t,i}$.

Our final method deviates from this tensor decomposition perspective by employing a deformation field $\D$ before querying $\mathcal{H}_i$ (see Equation~\ref{eq:canonical_feature_blending}). This effectively aligns spatial features in the hashtables across timesteps by explaining parts of the motion with the deformation field.

Another way of achieving a 4D tensor decomposition is presented in concurrent works by \citet{attal2023hyperreel, kplanes_2023, hex_planes}, who combine features from the 6 possible 2D feature planes instead of 4 outer-products between 1D and 3D tensors, as in Equation~\ref{eq:tensorf}.

\paragraph{Relation to HyperNeRF}
HyperNeRF~\cite{park2021hypernerf} adds so-called \emph{ambient dimensions} to their canonical space NeRFs to resolve topological issues that cannot be modeled by a deformation field. 
Instead of adding continuous ambient dimensions, our method models the canonical space with multiple hash grids, essentially introducing a \emph{discrete} ambient dimension that serves a similar purpose.

%% file: include/05_experimental_results.tex
\begin{figure*}
    \centering
    \includegraphics[width=\textwidth]{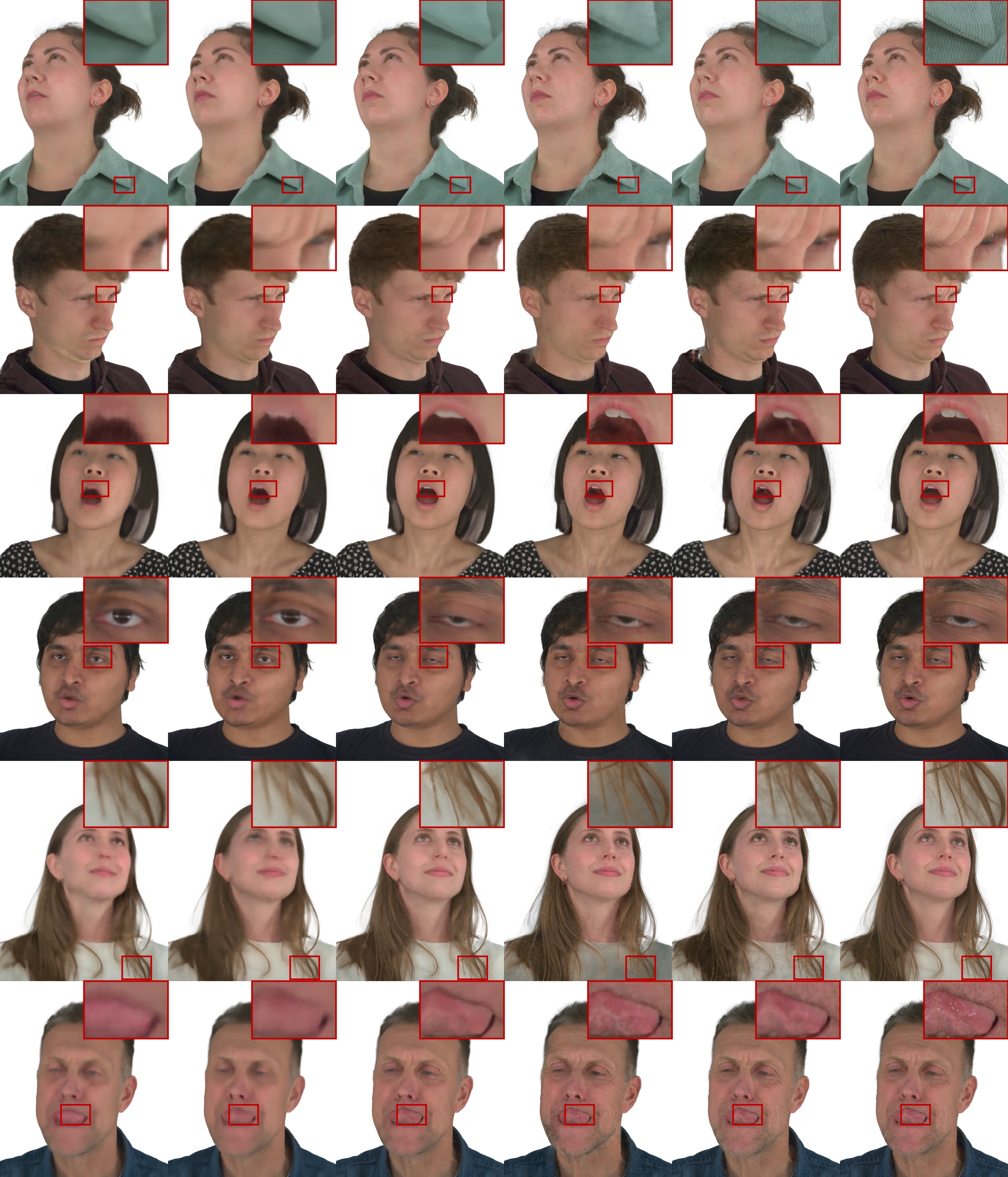}
    \begin{tabularx}{\textwidth}{CCCCCC}
        NeRFies & HyperNeRF & DyNeRF & Instant~NGP & Ours & GT \\
    \end{tabularx}
    \setlength{\abovecaptionskip}{-0.3cm}
    \caption{\textbf{Qualitative results.} Our method reconstructs high-quality detail even for challenging expressions.}
    \label{fig:06_results_comparison}
\end{figure*}

\section{Experimental Results}

We evaluate our method on the task of novel view synthesis (NVS) from multi-view video recordings on 10 diverse sequences from our dataset that focus on different aspects of facial and head movements. The validation sequences contain head rotations, laughs, eye blinking, talking, hair shaking, various mouth movements as well as one free expression sequence. All videos consist of 300-500 frames at 73~fps. 
We choose 12 out of the 16 available viewpoints as input and evaluate the NVS task on the remaining 4.
The selected evaluation views are equally distributed across the camera setup, resulting in a challenging evaluation protocol due to the presence of extreme viewing angles (see Figure~\ref{fig:05_evaluation_cam_ids}).

\subsection{Data Preparation}

Before running our experiments, we exploit the controlled nature of our dataset to facilitate reconstruction of the dynamic 3D scenes from image inputs. In concrete terms, we perform the following preprocessing steps:

\paragraph{Depth maps generation.} We employ the standard COLMAP pipeline to obtain depth maps for each of the 12 training views \cite{schoenberger2016mvs_colmap, schoenberger2016sfm_colmap}. To remove noisy depth measurements, we discard depth values observed by fewer than 3 cameras.

\paragraph{Background matting.} We use Background Matting v2~\cite{BGMv2} to obtain an alpha map given a captured frame and corresponding background image. To ensure the best quality, we use their ResNet101~\cite{he2016deep} version and set the error threshold to 0.01 in the refinement stage.

\paragraph{Image downsampling.} For all of our experiments, we downsample images by a factor of two to $1604 \times 1100$~pixels, which is sufficient for all methods. Temporally, we do not downsample and conduct all experiments on the full 73~fps.

\paragraph{Color correction.} Despite the color calibration of our cameras, there can still be slight differences in brightness across views. To address this, we first use facial segmentation masks~\cite{yu2018bisenet} to sample pixel values from the face, the torso, and the hair region. We then align the obtained color distributions across views by solving for an affine color transformation matrix using optimal transport \cite{flamary2021pot}.

\subsection{Floater Removal}
\label{sec:anti_floater}
Grid-based scene representations generally lack the induced smoothness prior of pure MLP architectures. As a result, they tend to generate small floaters that impair the visual quality of re-renderings. Since our hash ensemble is based on Instant~NGP, it inherits this tendency. To address this, we specify tight-fitting, axis-aligned bounding boxes for each sequence and only reconstruct radiance fields inside. In addition to tight scene box fitting, which we make available to all baselines, {\OURS} employs the following two techniques to suppress floaters, which are ablated in Section~\ref{sec:ablation_floater}.

\paragraph{View Frustum Culling.} We exclude regions in space that are seen by less than 2 train cameras and are thus especially prone to produce floaters. These regions are neither queried during training nor inference.

\paragraph{Occupancy Grid Filtering.} 
Before inference, we apply a low-pass filter to the density grid that our Instant~NGP backbone tracks during training and only render within the largest connected component, effectively discarding small isolated islands of density.

\subsection{Implementation Details}
\label{sec:implementation_details}
We implement our method in PyTorch~\cite{pytorch} within the Nerfstudio~\cite{nerfstudio} framework, which uses the NerfAcc~\cite{li2022nerfacc} implementation of Instant~NGP.

We train all our models for 300k iterations using a warmup schedule of $E_{\text{init}} = E_{\text{trans}} = 40$k, which takes approximately one day on a single Nvidia RTX A6000. The inference of a single frame at a resolution of $1604 \times 1100$ pixels takes roughly 25 seconds.

We use a learning rate of $1e^{-3}$ for all model components, which is decayed by a factor of $0.8$ every 20k iterations. For the deformation field $\D$, we use a factor of $0.5$ instead, such that the learning rate is sufficiently low after the warm-up phase.

Furthermore, we use $N=32$ hash tables, each configured with the default hyperparameters of \citet{mueller2022instant}. For our deformation field $\D$, we use the default configuration of the $SE(3)$ field by \cite{park2021nerfies} and 128 dimensions for the learnable deformation codes $\z_t$. Our blend weights $\beta \in \mathbb{R}^N$ have one weight per hash table.

\subsection{Baselines}
We compare our method against several state-of-the-art methods for NVS of dynamic scenes.
In particular, we compare against the following methods:

\subsubsection{Dynamic NeRFs}

\hfill 

\textit{Nerfies~\cite{park2021nerfies}} serves as representative for deformable NeRFs. 
We use the same implementation as for HyperNeRF, but without the ambient dimensions.
\\

\textit{HyperNeRF~\cite{park2021hypernerf}} extends Nerfies to address topological issues. Due to memory issues with their official implementation, we port their code to the Nerfstudio framework and carefully choose hyperparameters to match the performance of the official implementation.
\\

\textit{DyNeRF~\cite{li2022dynerf}}, in contrast, is not constrained to represent dynamic content using a deformation field, but directly conditions a NeRF on a time-dependent latent code. Since no public code is available, we implement DyNeRF in Nerfstudio and finetune it to our data distribution.

\subsubsection{Time-Agnostic Methods}
Furthermore, the multi-view nature of our dataset allows for 3D reconstructions on a per-frame basis. Hence, we consider two additional baseline methods that do not consider time. First, we apply Poisson Surface Reconstruction (PSR)~\cite{psr_poisson_surface_reconstruction} on the COLMAP point clouds.
Second, we run the official Instant~NGP ~\cite{mueller2022instant} on each frame separately.

\begin{figure}[tb]
    \centering
    \includegraphics[width=\linewidth]{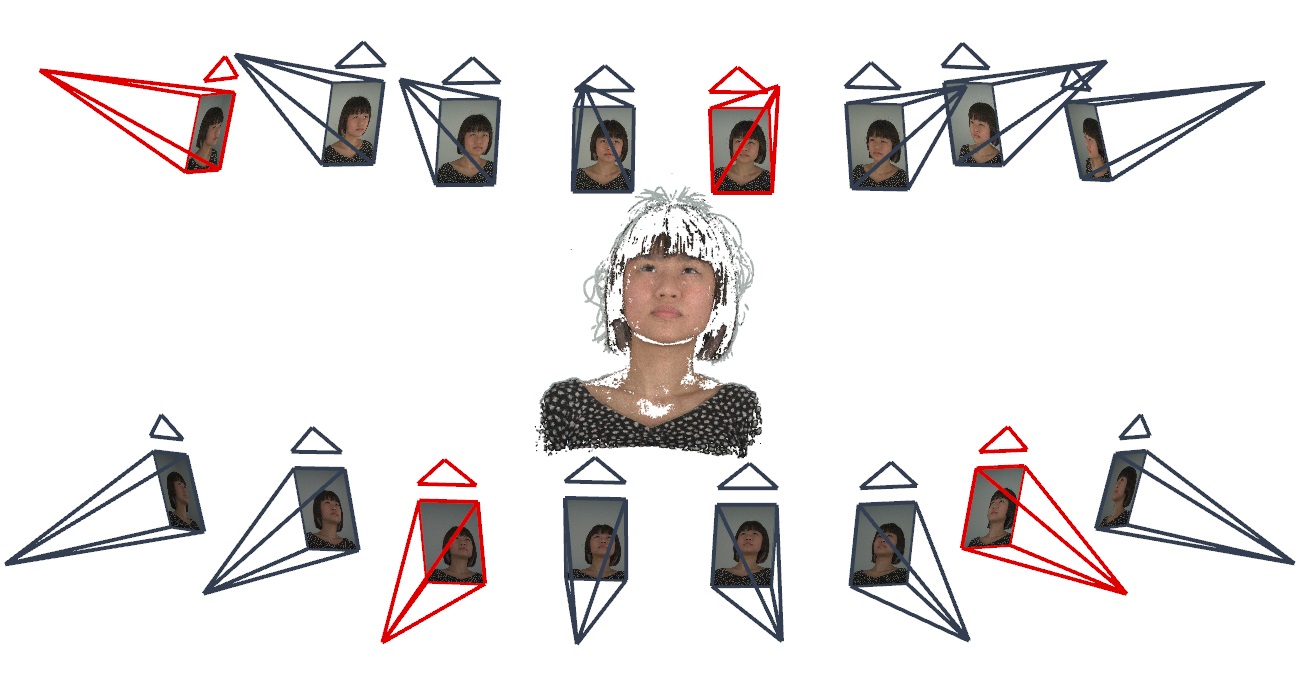}
    \caption{\textbf{Spatial layout of our camera setup.} Marked in red are the 4 views used for evaluation. The point cloud reconstruction is obtained via COLMAP and used for additional depth supervision.}
    \label{fig:05_evaluation_cam_ids}
\end{figure}

\subsubsection{Face-Specific Methods}
Additionally, we compare against Neural Head Avatars (NHA)~\cite{NHA} and NeRFace~\cite{Gafni_2021_CVPR_nerface} as representatives of face-specific dynamic reconstruction methods, both rely on the geometric prior provided by tracked statistic mesh models. 
NHA is a mesh-based method that optimizes for vertex offsets on top of the FLAME and predicts view- and expression-dependent textures.
NeRFace utilizes the 3DMM parameters directly to condition a NeRF and is thereby similar to DyNeRF.
Since both methods were initially designed for monocular use-cases, we expand them to our multi-view scenario by employing a custom multi-view FLAME tracker and providing all 12 views during training. 
Note, that the reliance on a 3DMM provides both methods with a certain degree of reanimation ability, but potentially impairs rendering quality when provided with dense enough observations.

\subsubsection{Background Modeling}
For a fair comparison, we encourage all NeRF-based baselines to represent the background without density, by coloring all remaining transmittance as white. For this purpose, we use our alpha masks to set all background pixels in the ground truth images to white as well. In our experience, this simple technique allows all baselines to learn good reconstructions of the person in the foreground.

\input{tables/main_evaluation}

\subsection{Evaluation Protocol}

We evaluate all methods on 4 held-out camera viewpoints. Figure~\ref{fig:05_evaluation_cam_ids} shows the spatial arrangement of the evaluation cameras. Furthermore, in the interest of compute time, we only evaluate the prediction on 15 evenly distributed timesteps from each evaluation camera. We verified on multiple sequences that all employed image metrics differ by at most 0.02 points when evaluating only 15 timesteps instead of the full sequence. 

\paragraph{Metrics}
We report three image metrics to evaluate the visual quality of individual reconstructed frames:
Peak Signal-to-Noise Ratio (PSNR), Structural Similarity (SSIM) \cite{wang2004image}, and Learned Perceptual Image Patch Similarity (LPIPS) \cite{zhang2018unreasonable}.
All metrics are evaluated on a per-frame basis after blending predictions with the alpha masks in order to focus on the facial region.
Additionally, we compute a JOD metric~\cite{jod_metric} used by \cite{li2022dynerf}, which indicates perceptual difference to a reference video.

\subsection{Comparison to State of the Art}

Table~\ref{tab:05_main_results} shows that {\OURS} quantitatively outperforms all baselines in all image metrics. In particular, our method shows strong improvements in the SSIM and LPIPS metrics that are sensitive to high-frequency details. This observation is matched by the qualitative comparison in Figure~\ref{fig:06_results_comparison}, where our method reconstructs better facial detail. We recommend the reader to watch the supplementary video for a more in-depth visual analysis of our method.

\begin{figure}
    \centering
    \begin{tabularx}{\linewidth}{l X}
        \rotatebox[origin=r]{90}{\parbox[c]{3.7cm}{\centering Instant NGP}} & \multirow{2}{*}{\includegraphics[width=\linewidth]{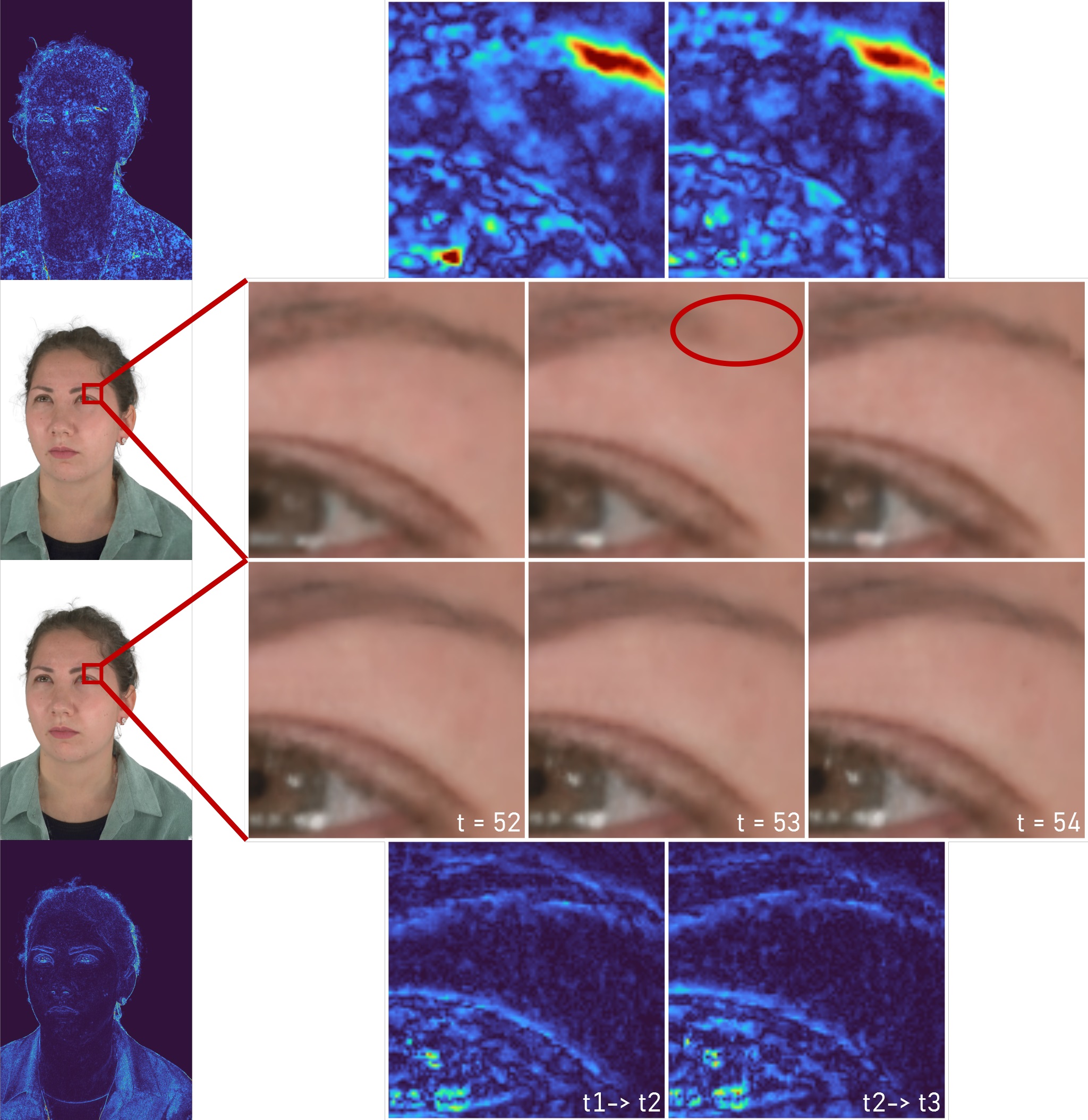}}\\[3.7cm]
        \rotatebox[origin=c]{90}{\parbox[c]{3.7cm}{\centering Ours}} & \\
    \end{tabularx}
    
    \caption{\textbf{Temporal consistency.} We show a re-rendering and its temporal difference image for a novel view (left). On the right side, we demonstrate the flickering artifact of the Instant~NGP baseline between three adjacent frames, where an eyebrow shrinks and grows between frames. In comparison, {\OURS} offers more temporal consistency.}

    \label{fig:07_temporal_consistency}
\end{figure}

\paragraph{Evaluation of Temporal Consistency}
\input{tables/jod_results}
Per-frame metrics such as PSNR, SSIM, and LPIPS do not account for temporal artifacts such as flickering. Hence, we employ the perceptual video metric JOD~\cite{jod_metric} to measure visual similarity of a rendered video to its ground truth counterpart. For all major baselines, we render videos at a third of the training framerate, i.e. 24.3 fps, and average the JOD scores over all validation views and all 10 validation sequences. The results of this temporal evaluation are given in Table~\ref{tab:jod_results}. Note, that the Instant~NGP baseline is completely time-agnostic, which leads to considerable flickering artifacts in video renderings. Figure~\ref{fig:07_temporal_consistency} shows an example of such an artifact. In contrast, {\OURS} provides a smooth temporal experience. 

\paragraph{Comparison to Instant~NGP}
The Instant~NGP baseline produces compelling images on a per-frame basis as can be seen in Figure~\ref{fig:06_results_comparison}. However, it suffers from a strong tendency to generate floaters and scattered surfaces due to the sparse nature of our camera setup. In contrast, {\OURS} constrains the space across multiple timesteps which greatly contributes towards removing floaters. This also holds for a {\OURS} trained without any anti-floater strategies or additional losses. Such a bare-bones version of our model still outperforms Instant~NGP (see the top row in Table~\ref{tab:floater_removal_techniques}). This shows that in our sparse setting, having higher expressiveness by modeling each frame independently (e.g., the Instant~NGP baseline has 10-15 times more parameters than our model) does not lead to better reconstructions.

\input{tables/face_specific_methods}
\input{tables/dynerf_dataset}

\paragraph{Comparison to Face-Specific Methods}
To compare against NHA and NeRFace, we evaluate on 7 sequences from our dataset, excluding 3 with more complex motion where the preprocessing pipeline of NHA failed to predict facial landmarks, segmentation masks and normals. Furthermore, NHA only synthesizes the head without a torso. Therefore, we only evaluate the facial region for a fair comparison. Table~\ref{tab:face_specific_methods} shows that {\OURS} outperforms both baselines despite them being specifically designed for faces.

\begin{figure*}
    \centering
    \includegraphics[width=\textwidth]{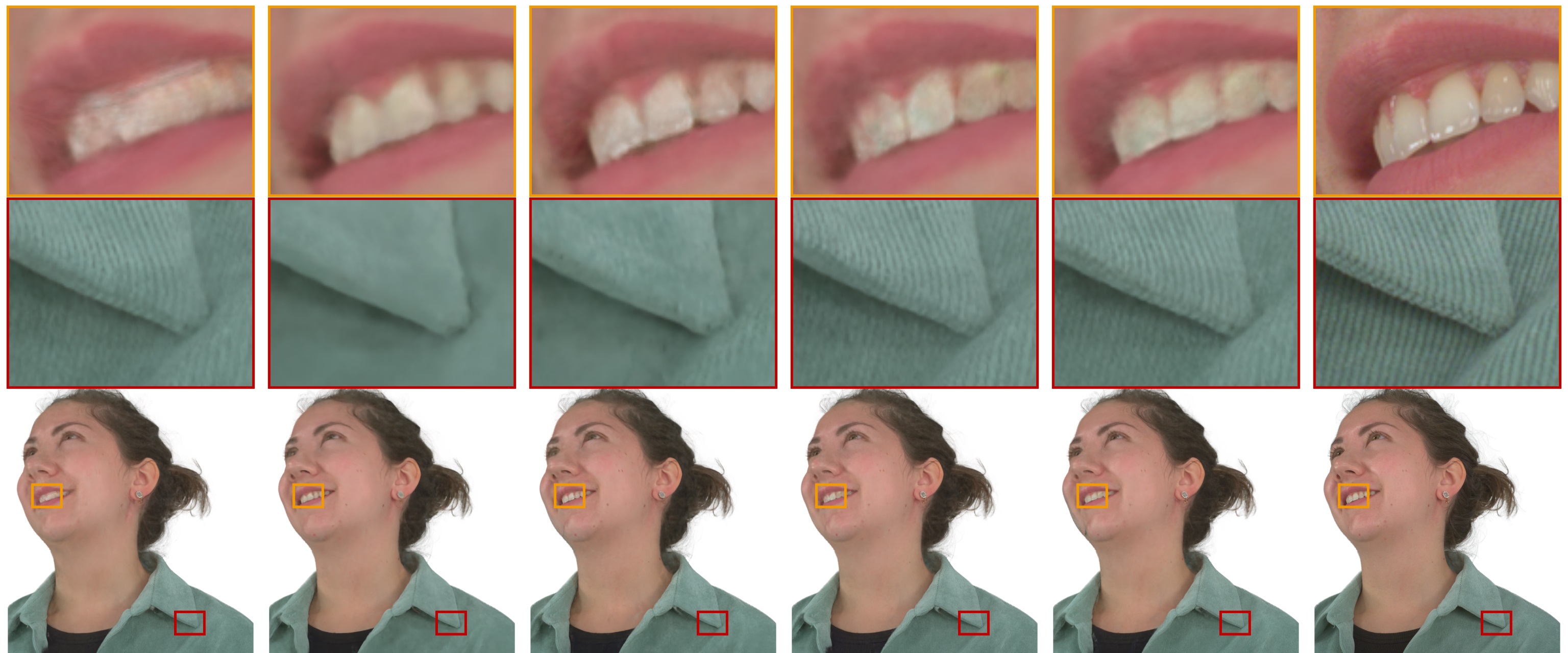}
    \begin{tabularx}{\textwidth}{CCCCCC}
        (a) NGP + Def. & (b) Hash Ensemble & (c) w/o Warmup & (d) 16 Hash Tables & (e) Ours & (f) GT\\
    \end{tabularx}
    \setlength{\abovecaptionskip}{-0.3cm}
    \caption{\textbf{Ablation of our model components.} Combining Instant NGP with a deformation field (a) produces sharp detail in rigidly moving areas of the scene, e.g., the torso, but struggles with more challenging motion such as mouth movements. On the other hand, employing an ensemble of hash encodings (b) can better deal with complex motions but generally produces more blurry reconstructions. Combining both components (c) leverages the strength of both architectures but still does not produce the same detail in rigidly moving areas as an Instant NGP with deformation field. By employing a warmup phase, sharp detail already returns when 16 tables are used (d) which can be further improved by increasing the number of hash encodings (e).}
    \label{fig:08_ablation_comparison}
\end{figure*}

\paragraph{Comparison on Neural 3D Video Dataset~\cite{li2022dynerf}}

{\OURS} does not make strong assumptions on the content of a dynamic scene and is therefore applicable to more general scenarios. To demonstrate {\OURS}'s generality, we evaluate on the 6 publicly available sequences of the Neural 3D Video dataset~\cite{li2022dynerf}. We follow the evaluation protocol of NeRFPlayer~\cite{song2022nerfplayer} and HyperReel~\cite{attal2023hyperreel} that downsample the recordings to 1352 x 1014 pixel resolution, hold out the top central view for evaluation, and report metrics averaged over all 6 sequences. We further re-compute the poses with COLMAP~\cite{schoenberger2016mvs_colmap} as the ones provided with the dataset are slightly off. Table~\ref{tab:dynerf_dataset_results} shows the quantitative results. We excluded the original DyNeRF~\cite{li2022dynerf} as well as StreamRF~\cite{streamrf} from the evaluation as their numbers were only computed on 1 of the 6 available sequences and are thus not comparable to the results of NeRFPlayer and HyperReel. The evaluation shows that NeRSemble can reasonably model generic dynamic scenes despite its functionality being inspired by facial blendshapes. However, since our method relies on Instant NGP, it also inherits some of its weaknesses. In particular, it does not model light refraction as HyperReel does. As a result, {\OURS} cannot perfectly capture the effects of window panes and glass bottles which are prevalent in the Neural 3D Video dataset.

\subsection{Ablations}

In addition to the comparison against baselines, we conduct several experiments to validate our design choices and understand the inner workings of \OURS.

\paragraph{Contribution of Architectural Components} 
We ablate the effect of using a hash ensemble and the deformation field. Table~\ref{tab:05_main_results} shows that neither a deformation field with an Instant~NGP backbone (NGP + Def.) nor a plain hash ensemble matches the performance of our final model. However, both architectures are strong baselines on their own. In Figure~\ref{fig:08_ablation_comparison}, we present qualitative results, which show that the deformation-based approach generally produces sharper reconstructions, but struggles with more challenging motions that are difficult to model with deformations. On the other hand, the hash ensemble has the expressiveness to model any dynamic scene via feature blending but will typically produce more blurry results for simple movements, since it is missing the prior of a deformation field.
The quantitative results in Table~\ref{tab:05_main_results} confirm these findings, with the hash ensemble scoring a high PSNR but worse LPIPS value.

\paragraph{Number of Hash Tables} \OURS~ with 16 hash tables only suffers a negligible amount compared to 32 hash tables. This confirms that the ratio between the number of frames and hash tables scales well and information is shared effectively across tables.

\input{tables/floater_removal_strategies}
\paragraph{Effect of Warm-Up Phase} Training without warm-up consistently performs worse. We attribute this to the fact that giving the model access to all hash grids right away prevents it from learning correspondences with the deformation field. As a result, the learned hash encodings are less well-aligned and cannot be blended as effectively. Visually, this manifests in slightly blurrier renderings. This insight is in line with HyperNeRF's proposal to disable the use of ambient dimensions in the beginning.

\paragraph{Effect of Depth Supervision} Since removing the depth supervision only slightly impairs the performance, we hypothesize that the RGB information of the 12 input views already sufficiently supervises the geometry. However, exploiting depth supervision from orthogonal channels, such as a fitted 3DMM or a trained depth prediction network, could still be beneficial as it incorporates data priors from  sources other than the RGB video frames.

\paragraph{Floater Removal Techniques}
\label{sec:ablation_floater}
We ablate the effect of three strategies to suppress floaters. First, we isolate the effect of all additional losses, i.e. mask loss, depth supervision, and distortion loss, and note their significant impact on performance in Table~\ref{tab:floater_removal_techniques}. View frustum culling and occupancy grid filtering, on the other hand, do not affect the reported metrics but still improve visual quality when rendering novel camera trajectories.

\paragraph{Content of Individual Hash Grids}

\begin{figure}[t]
    \centering
    \includegraphics[width=\linewidth]{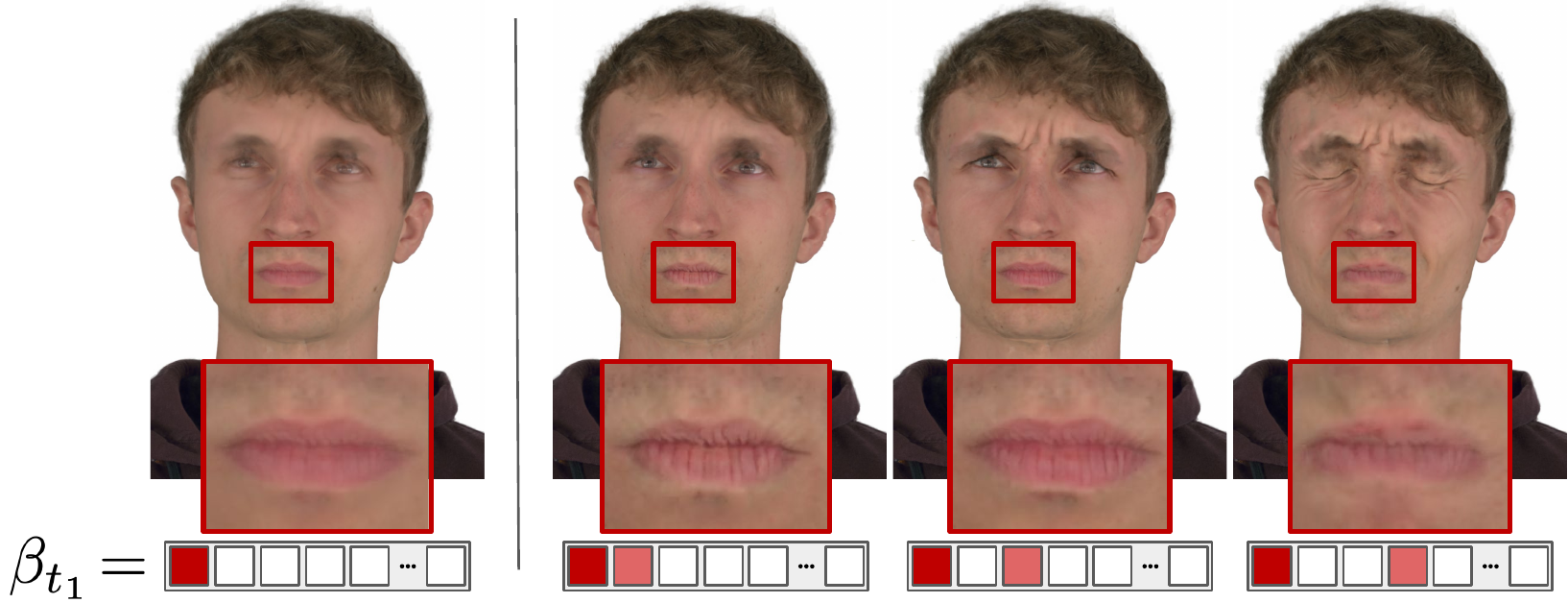}
    \caption{ \textbf{Blend weights.}
    Investigating the contents of the first hash grid by setting $\beta_{t_1, i} = 0~(i >1)$ reveals that the first hash grid stores some sort of average representation (left). On the right we successively set $\beta_{t_1, i} = 0.75$ for $i \in \{2, 3, 4\}$. Each table stores additional details that are exceeding the representational capacity of the deformation network. Note that we use $\omega_{t_1}$ for all shown examples and $t_1$ denotes the first frame.
    }
    \label{fig:blend_weight_interpolation}
\end{figure}

We analyze the contents of the individual hash grids $\mathcal{H}_i$ in Figure \ref{fig:blend_weight_interpolation}. For this purpose, we modify the learned blend weights $\beta_{t_1, i}~ (i > 1)$ for the first frame $t_1$ of a sequence, while keeping $\beta_{t_1, 1}$ and the deformation codes $\omega_{t_1}$ fixed. This experiment reveals that the deformation field $\mathcal{D}$ accounts for rigid movements of the scene, since modifying $\beta_{t_1}$ results in well-aligned appearance changes while the head stays static. Furthermore, $\mathcal{H}_1$ seems to store a representation comparable to the mean face of the person. The remaining hash grids then behave similarly to a dynamic, volumetric texture that further adds details to the scene that are otherwise unexplained, e.g., topologically complicated deformations, expressions-dependent wrinkles, or illumination changes. We attribute the special status of the first hash grid $\mathcal{H}_1$ to the fact that it is always active during training while all other hash grids are gradually introduced during the warmup phase.

%% file: tables/main_evaluation.tex
\begin{table}[tb]
    \centering

    \caption{\textbf{Quantitative evaluation.} We perform comparisons against two non-temporal baselines as well as three dynamic reconstruction methods. We evaluate unseen validation views of 10 diverse sequences from our dataset. Our method outperforms the baselines in all three metrics. The bottom two rows show ablations of our method with respect to core architectural components and the training procedure.}
    
    \begin{tabular}{ll rrrrr}
        \toprule

            &  Method
                & {PSNR} $\uparrow$ & {SSIM} $\uparrow$ & {LPIPS} $\downarrow$
        \\
        
        \midrule

        \multirow{2}{*}{\rotatebox[origin=r]{90}{\parbox[c]{0.4cm}{\centering \small Static}}}
        
        & PSR
            & 12.5 & 0.774 & 0.341 \\

        & Instant NGP
            & 28.8 & 0.864 & 0.254 \\

        \midrule

        \multirow{3}{*}{\rotatebox[origin=r]{90}{\parbox[c]{0.8cm}{\centering \small Dynamic}}}

        & Nerfies
            & 29.5 & 0.849 & 0.299 \\

        & HyperNeRF
            & 29.6 & 0.848 & 0.304 \\

        & DyNeRF
            & 30.6 & 0.860 & 0.254 \\

        \midrule

        & \textbf{Ours}
            & \textbf{31.8} & \textbf{0.875} & \textbf{0.212} \\

        \midrule

        \multirow{2}{*}{\rotatebox[origin=r]{90}{\parbox[c]{0.4cm}{\centering \small Parts}}}
        
        & NGP + Def.
            & 30.8 & 0.864 & 0.231 \\

        & Hash Ensemble
            & 30.5 & 0.857 & 0.257 \\

        \midrule

        \multirow{3}{*}{\rotatebox[origin=r]{90}{\parbox[c]{0.8cm}{\centering \small Ablation}}}

        & w/o Depth
            & 31.5 & 0.873 & 0.217 \\

        & w/o Warmup
            & 31.0 & 0.866 & 0.234 \\

        & only 16 tables
            & 31.5 & 0.871 & 0.218 \\
            
        \bottomrule
    \end{tabular}

    \label{tab:05_main_results}
\end{table}

%% file: tables/jod_results.tex
\begin{table}[tb]
    \centering

    \caption{\textbf{Evaluation of temporal consistency} using the perceptual quality metric Just-Objectionable-Difference (JOD)~\cite{jod_metric}. Higher numbers indicate less temporal flickering and a greater resemblance to the ground truth video.}
    
    \begin{tabularx}{\linewidth}{Xccccc}
        \toprule
        
        Method
            & Inst. NGP
            & Nerfies
            & HyperNeRF
            & DyNeRF
            & Ours \\
                
        \midrule
            
        JOD $\uparrow$
            & 6.75 & 7.23 & 7.27 & 7.69 & \textbf{7.86} \\
    
        \bottomrule
    \end{tabularx}

    \label{tab:jod_results}
\end{table}

%% file: tables/face_specific_methods.tex
\begin{table}[tb]
    \centering

    \caption{\textbf{Evaluation against face-specific methods.} {\OURS} compares favorably to Neural Head Avatars~\cite{NHA} and NeRFace~\cite{Gafni_2021_CVPR_nerface}.}
    
    \begin{tabular}{lrrr}
        \toprule
        
            Method
                & {PSNR} $\uparrow$ & {SSIM} $\uparrow$ & {LPIPS} $\downarrow$
        \\
        
        \midrule
        Neural Head Avatars & 31.0 & 0.927 & 0.041 \\
        NeRFace & 35.2 & 0.956 & 0.047 \\
        Ours & \textbf{37.5} & \textbf{0.968} & \textbf{0.023} \\
            
        \bottomrule
    \end{tabular}

    \label{tab:face_specific_methods}
\end{table}

%% file: tables/dynerf_dataset.tex
\begin{table}[tb]
    \caption{\textbf{Quantitative comparison on the Neural 3D Video Dataset.} Although {\OURS}'s functionality is inspired by facial blendshapes, it can also reasonably model generic dynamic scenes.}

    \centering
    \begin{tabular}{lcccc}
        \toprule
         Method 
            & \makecell[c]{NeRFPlayer \\\small(Instant NGP)}
            & \makecell[c]{NeRFPlayer \\\small(TensoRF)}
            & HyperReel 
            & Ours \\
         \midrule
         PSNR $\uparrow$ 
            & 30.3 
            & 30.7
            & \textbf{31.1} 
            & 29.9 \\
         \bottomrule
    \end{tabular}
    
    \label{tab:dynerf_dataset_results}
\end{table}

%% file: tables/floater_removal_strategies.tex
\newcommand{\checkbox}{\scalebox{1.5}{$\boxtimes$}}
\newcommand{\emptybox}{\scalebox{1.5}{$\square$}}

\begin{table}[tb]

    \caption{\textbf{Ablation of floater removal techniques.} Both view frustum culling~(VFC) and occupancy grid filtering~(OGF) have a negligible effect on the metrics as they mostly remove floaters in areas that are omitted in our evaluation protocol. Note that a plain version of {\OURS} without any additional losses~($\mathcal{L}$) or floater removal techniques already performs competitively compared to all baselines in Table~\ref{tab:05_main_results}.}
    
    \centering
    \begin{tabular}{ccc rrr}
        \toprule
        
            $\mathcal{L}$ & VFC & OGF
            & {PSNR} $\uparrow$ & {SSIM} $\uparrow$ & {LPIPS} $\downarrow$
        \\
        
        \midrule

        \emptybox & \emptybox & \emptybox
            & 30.4 & 0.868 & 0.230 \\
        \checkbox & \emptybox & \emptybox
            & 31.8 & \textbf{0.875} & 0.213 \\
        \checkbox & \checkbox & \emptybox
            & 31.8 & \textbf{0.875} & 0.213 \\
        \checkbox & \emptybox & \checkbox
            & \textbf{31.9} & \textbf{0.875} & \textbf{0.212} \\
        \checkbox & \checkbox & \checkbox
            & 31.8 & \textbf{0.875} & \textbf{0.212} \\
            
        \bottomrule
    \end{tabular}
    
    \label{tab:floater_removal_techniques}

\end{table}

%% file: include/06_conclusion.tex
\subsection{Limitations}

\begin{figure}
    \centering
    \begin{subfigure}[b]{0.04\linewidth}
        \begin{tabularx}{\linewidth}{l}
            \rotatebox[origin=r]{90}{\parbox[t]{1.2cm}{\centering {\small Prediction}}}\\
            \rotatebox[origin=c]{90}{\parbox[b]{1.15cm}{\centering {\small GT}}}
        \end{tabularx} 
        \caption*{}  %
    \end{subfigure}%
    \hfill%
    \begin{subfigure}[b]{0.37\linewidth}
        
         \centering
         \includegraphics[width=\linewidth]{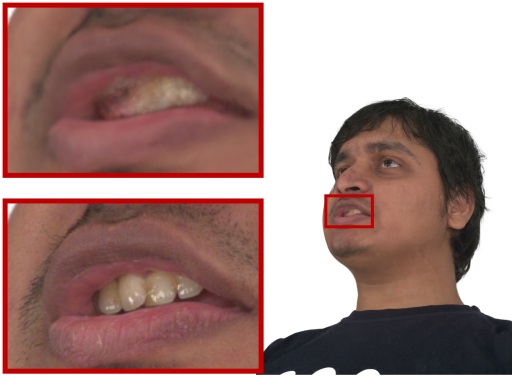}
         
         \caption{}
         \label{fig:failure_case_mouth}
    \end{subfigure}%
    \hfill%
    \begin{subfigure}[b]{0.37\linewidth}
         \centering
         \includegraphics[width=\linewidth]{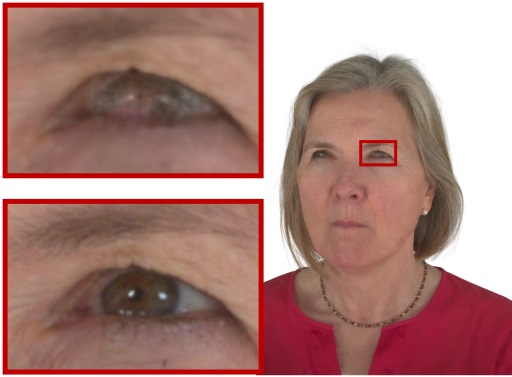}
         
         \caption{}
         \label{fig:failure_case_eye}
    \end{subfigure}%
    \hfill %
    \begin{subfigure}[b]{0.18\linewidth}
         \centering
         \includegraphics[width=\linewidth]{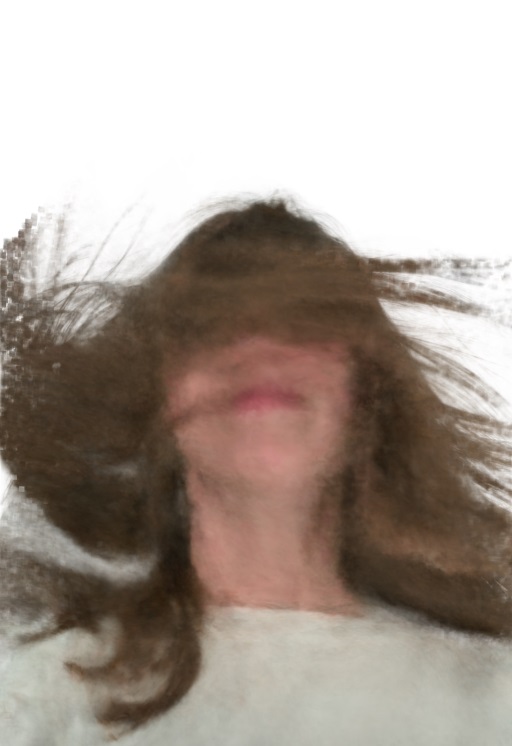}
         \caption{}
         \label{fig:failure_case_hair}
    \end{subfigure}
    \hfill
    \caption{\textbf{Failure cases.} The high degree of occlusion of the mouth interior can sometimes cause a hollow face illusion where teeth are falsely reconstructed at the back of the mouth~(a). Specular reflections of the light sources in the eyes may cause rare eye artifacts~(b). The deformation field may fail to model extremely fast hair motion, which hinders the canonical hash grids from synthesizing a sharp result for some frames~(c).}
    \label{fig:failure_cases}
\end{figure}

In our experiments, we demonstrate that we can achieve convincing results with a sparse set of multi-view recordings; however, various limitations remain. 
Since {\OURS} models explicit correspondences across timesteps via a deformation field, it cannot perfectly capture fast hair motion (see Figure~\ref{fig:failure_case_hair}). %
To address this, incorporating movement priors via optical flow or differentiable physics could be an interesting field for future work. \\
Furthermore, our method currently focuses on recovering the appearance and motion of a specific sequence by optimizing for the dynamic radiance field representation.
As a result, our method is unable to learn priors that generalize across sequences. 
Here, we see great potential for future work on dynamic NeRFs that generalize over both identities and facial expressions. 
A learned prior over the distribution of realistic 4D avatars could help to further constrain the optimization procedure. 
This would be particularly important for monocular inputs or capturing facial regions, such as the mouth interior, that are often occluded during the majority of recording time and may thus exhibit inferior reconstruction quality (see Figure~\ref{fig:failure_cases}).

\section{Conclusion}

In this work, we have proposed a new method and dataset focusing on the radiance field reconstruction of animated human heads from multi-view video inputs.
To this end, we have introduced a novel multi-view video benchmark of diverse human heads containing over {\nroughparticipants} identities with {\nroughsequences} sequences.
We further proposed a new method for generating photo-realistic re-renderings of arbitrary viewpoints and time steps, and hope that our dataset and accompanying benchmark will be an important contribution to the community, and facilitate future work on digital humans.

Our proposed novel representation for spatio-temporal NeRFs uses deformation fields to factor out coarse movements and an ensemble of hash grid encodings to model fine deformations and increase the temporal capacity of our model.
Our experiments demonstrate that \OURS~ achieves temporally coherent and highly detailed volumetric reconstructions from multi-view video inputs, outperforming existing baselines by a significant margin, in particular when sequences contain complex motions.